\documentclass[11pt,a4paper]{article}
\usepackage[hyperref]{naaclhlt2018}
\usepackage{times}
\usepackage{latexsym}

\usepackage[utf8]{inputenc}
\usepackage[LGR,T1,OT2,OT1]{fontenc}
\newcommand\cyr
{
\renewcommand\rmdefault{wncyr}
\renewcommand\sfdefault{wncyss}
\renewcommand\encodingdefault{OT2}
\normalfont
\selectfont
}
\DeclareTextFontCommand{\textcyr}{\cyr}

\newcommand{\textgreek}[1]{\begingroup\fontencoding{LGR}\selectfont#1\endgroup}
\usepackage{CJKutf8}

\usepackage{url}

\usepackage{arydshln}
\usepackage{graphvizzz}
\usepackage{amsmath}
\usepackage{multirow}
\usepackage{url}
\usepackage{stmaryrd}
\usepackage{qtree}
\usepackage{tikz-qtree}
\usepackage{graphicx}
\usetikzlibrary{matrix}
\usetikzlibrary{fadings}
\usetikzlibrary{chains}
\usetikzlibrary{snakes}
\usetikzlibrary{arrows}
\usetikzlibrary{shapes}
\usetikzlibrary{automata}
\usetikzlibrary{decorations.pathreplacing,shapes,backgrounds}
\usetikzlibrary{fit}
\usepackage{array}
\usepackage{alltt}
\usepackage{listings}
\DeclareMathOperator*{\argmax}{arg\,max}

\usetikzlibrary{positioning,automata}
\usepackage{CJK,CJKspace,CJKpunct}
\usepackage{array}
\usepackage{tabulary}
\newcolumntype{K}[1]{>{\centering\arraybackslash}p{#1}}

\usepackage{algorithm}
\usepackage[noend]{algpseudocode}
\usepackage{pifont}
\usepackage{etoolbox}\AtBeginEnvironment{algorithmic}{\footnotesize}

\usepackage{color, colortbl}
\definecolor{Gray}{gray}{0.9}
\definecolor{darkolivegreen}{rgb}{0.33, 0.42, 0.18}
\definecolor{darkseagreen}{rgb}{0.56, 0.74, 0.56}
\usepackage{amssymb}

\newcolumntype{P}[1]{>{\raggedright\arraybackslash}p{#1}}

\aclfinalcopy 

\title{Polyglot Semantic Parsing in APIs}

\author{Kyle Richardson$^{\dag}$, Jonathan Berant$^{\ddag}$, Jonas Kuhn${^\dag}$ \\
  $^{\dag}$Institute for Natural Language Processing, University of Stuttgart, Germany  \\
  {\tt \{kyle,jonas\}@ims.uni-stuttgart.de} \\ 
  $^{\ddag}$Tel-Aviv University, Israel  \\
  {\tt joberant@cs.tau.ac.il} \\
  \\}

\date{}

\begin{document}
\maketitle
\begin{abstract}   
%
Traditional approaches to semantic parsing (SP) work by training individual models for each available parallel dataset of text-meaning pairs. In this paper, we explore the idea of polyglot semantic translation, or learning semantic parsing models that are trained on multiple datasets and natural languages.  In particular, we focus on translating text to code signature representations using the software component datasets of \newcite{richardson:17b,richardson:17a}.  The advantage of such models is that they can be used for parsing  a wide variety of input natural languages and output programming languages, or mixed input languages, using a single unified model. To facilitate modeling of this type, we develop a novel graph-based decoding framework that achieves state-of-the-art performance on the above datasets, and apply this method to two other benchmark SP tasks. 

\end{abstract}

\section{Introduction}

Recent work by \newcite{richardson:17b,richardson:17a,pydocs:17} considers the problem of translating source code documentation to lower-level code template representations as part of an effort to model the meaning of such documentation. Example documentation for a number of programming languages is shown in Figure 1, where each \emph{docstring} description in red describes a given function (blue) in the library. While capturing the semantics of docstrings is in general a difficult task, learning the translation from descriptions to formal code representations  (e.g., formal representations of functions) is proposed as a reasonable first step towards learning more  general natural language understanding models in the software domain. Under this approach, one can view a software library, or API, as a kind of parallel translation corpus for studying $\emph{text} \to \texttt{code}$ or $\texttt{code} \to \emph{text}$ translation. 








\begin{figure}[t]
\noindent\fbox{%
    \parbox{7.5cm}{%
\begin{centering}    
{\footnotesize \text{1. ($en$, \texttt{Java}) Documentation    }}  \\[-.7cm]    
\end{centering}
     {\scriptsize \begin{alltt}\textcolor{darkolivegreen}{
%/* \\     
*\textcolor{red}{\textbf{Returns the greater of two long values}} \\
%* \\
%*/ @see java.lang.Long\#MAX\_VALUE \\ 
%*/ ... \\ 
}\text{\textcolor{blue}{public static} long max(long a, long b)} \\[-.4cm]
    \end{alltt}
    }}}
\\[.1cm]
\noindent\fbox{%
    \parbox{7.5cm}{%
\begin{centering}    
{\footnotesize \text{2. ($en$, \texttt{Python}) Documentation}} \\[-.7cm]    
\end{centering}
     {\scriptsize \begin{alltt}
%\textcolor{gray}{\# from decimal.Context} \\[.2cm]
\textcolor{blue}{max}(self, a, b): \\
\hspace*{.2in}\textbf{\textcolor{red}{"""Compares two values numerically}} \\
\hspace*{.2in}\textbf{\textcolor{red}{and returns the maximum"""}} \\[-.4cm]
   \end{alltt}}}}
\\[.1cm] 
\noindent\fbox{%
    \parbox{7.5cm}{%
\begin{centering}    
{\footnotesize \text{3. ($en$, \texttt{Haskell}) Documentation}}  \\[-.7cm]    
\end{centering}
     {\scriptsize \begin{alltt}
--| \textbf{\textcolor{red}{"The largest element of a non-empty structure"}} \\     
\textcolor{blue}{maximum} :: forall z. Ord a a => t a -> a \\[-.4cm]
   \end{alltt}}}}
   \\[.1cm]
\noindent\fbox{%
    \parbox{7.5cm}{%
\begin{centering}    
{\footnotesize \text{4. ($de$, \texttt{PHP}) Documentation}}  \\[-.7cm]    
\end{centering}
     {\scriptsize \begin{alltt}\textcolor{darkolivegreen}{
%/* \\     
*\textcolor{red}{\textbf{gibt den gr{\"o}{\ss}eren dieser Werte zur{\"u}ck.}} \\
%* \\
%*/ @see java.lang.Long\#MAX\_VALUE \\ 
%*/ ... \\ 
}\text{\textcolor{blue}{max} (mixed \$value1, mixed \$value2)} \\[-.4cm]
    \end{alltt}
    }}}
\caption{Example source code documentation.}
\end{figure}


\newcite{richardson:17a} extracted the standard library documentation for 10 popular programming languages across a number of natural languages to study the problem of text to function signature translation. Initially, these datasets were proposed as a resource for studying semantic parser induction \cite{mooney2007learning}, or for building  models that learn to translate text to formal meaning representations from parallel data. In follow-up work \cite{richardson:17b}, they proposed using the resulting models to do automated question-answering (QA) and code retrieval on target APIs, and experimented with an additional set of software datasets built from 27 open-source Python projects.

As traditionally done in SP \cite{zettlemoyer2012learning}, their approach involves learning individual models for each parallel dataset or language pair, e.g., ($en$,\texttt{ Java}), ($de$,\texttt{ PHP}), and ($en$, \texttt{ Haskell}). Looking again at Figure 1, we notice that while programming languages differ in terms of representation conventions, there is often overlap between the functionality implemented and naming in these different languages (e.g., the \texttt{max} function), and redundancy in the associated linguistic descriptions.  In addition, each English description (Figure 1.1-1.3) describes \texttt{max} differently using the synonyms \emph{greater, maximum, largest}.  In this case, it would seem that training models on multiple datasets, as opposed to single language pairs, might make learning more robust, and help to capture various linguistic alternatives. 

With the software QA application in mind, an additional limitation is that their approach does not allow one to freely translate a given description to multiple output languages, which would be useful for comparing how different programming languages represent the same functionality. The model also cannot translate between natural languages and programming languages that are not observed during training. While software documentation is easy to find in bulk, if a particular API is not already documented in a language other than English (e.g., \texttt{Haskell} in $de$), it  is unlikely that such a translation will appear without considerable effort by experienced translators. Similarly, many individual APIs may be too small or poorly documented to build individual models or QA applications, and will in some way need to bootstrap off of more general models or resources.  

To deal with these issues, we aim to learn more general text-to-code translation models that are trained on multiple datasets simultaneously. Our ultimate goal is to build \emph{polyglot} translation models (cf. \newcite{johnson2016google}), or models with shared representations that can translate any input text to any output programming language, regardless of whether such language pairs were encountered explicitly during training. Inherent in this task is the challenge of building an efficient polyglot decoder,  or a translation mechanism that allows such crossing between input and output languages. A key challenge is ensuring that such a decoder generates well-formed code representations, which is not guaranteed when one simply applies standard decoding strategies from SMT and neural MT (cf. \newcite{cheng2017learning}). Given our ultimate interest in API QA, such a decoder must also facilitate monolingual translation, or being able to translate to specific output languages as needed.  

To solve the decoding problem, we introduce a new graph-based decoding and representation framework that reduces to solving shortest path problems in directed graphs. We investigate several translation models that work within this framework, including traditional SMT models and models based on neural networks, and report state-of-the-art results on the technical documentation task of \newcite{richardson:17a,richardson:17b}. To show the applicability of our approach to more conventional SP tasks, we apply our methods to the GeoQuery domain \cite{zelle} and the Sportscaster corpus \cite{chenArticle}. These experiments also provide insight into the main technical documentation task and highlight the strengths and weaknesses of the various translation models being investigated. 

\section{Related Work}
Our approach builds on the baseline models introduced in \newcite{richardson:17a} (see also \newcite{deng}). Their work is positioned within the broader SP literature, where traditionally  SMT \cite{wongMAIN} and parsing \cite{ZettlemoyerO} methods are used to study the problem of translating text to formal meaning representations, usually centering around QA applications \cite{berant2013semantic}. More recently, there has been interest in using neural network approaches  either in place of \cite{dong2016language,kovcisky2016semantic} or in combination with \cite{misra2016neural,jia2016data,cheng2017learning} these traditional models, the latter idea we look at in this paper. 

Work in NLP on software documentation has accelerated in recent years due in large part to the availability of new data resources through websites such as StackOverflow and Github (cf. \newcite{allamanis2017survey}). Most of this recent work focuses on processing large amounts of API data in bulk \cite{gu2016deep,pydocs:17}, either for learning longer  executable programs from text  \cite{yin2017syntactic,rabinovich2017abstract}, or solving the inverse problem of code to text generation \cite{iyer2016summarizing,richardson2017code2text}. In contrast to our work, these studies do not look explicitly at translating to target APIs, or at non-English documentation. 

  
The idea of polyglot modeling has gained some traction in recent years for a variety of problems \cite{tsvetkov2016polyglot} and has appeared within work in SP under the heading of \emph{multilingual SP} \cite{jie2014multilingual,duong2017multilingual}.  A related topic is learning from multiple knowledge sources or domains \cite{herzig2017neural}, which is related to our idea of learning from multiple APIs. When building models that can translate between unobserved language pairs, we use the term \emph{zero-shot translation} from \newcite{johnson2016google}. 

\section{Baseline Semantic Translator}


\paragraph{Problem Formulation}Throughout the paper, we refer to target code representations as API \emph{components}. In all cases, components will consist of formal representations of functions, or function signatures (e.g., \texttt{long max(int a, int b)}), which include a function name (\texttt{max}), a sequence of arguments (\texttt{int a,  int b}), and other information such as a return value (\texttt{long}) and namespace (for more details, see \newcite{richardson2018language}).  For a given API dataset $D = \{(\textbf{x}_{i},\textbf{z}_{i})\}_{i=1}^{n}$ of size $n$,  the goal is to learn a model that can generate \emph{exactly} a correct component sequence $\textbf{z} = (z_{1},..,z_{| \textbf{z} |})$, within a finite space $\mathcal{C}$ of signatures (i.e., the space of all defined functions), for each input text sequence $\textbf{x} = (x_{1},...,x_{| \textbf{x} |})$. This involves learning a probability distribution $p(\textbf{z} \mid \textbf{x})$. As such, one can think of this underlying problem as a \emph{constrained} MT task. 

In this section, we describe the baseline approach of \newcite{richardson:17a}. Technically, their approach has two components: a simple word-based translation model and task specific decoder, which is used to generate a $k$-best list of candidate component representations for a given input $\textbf{x}$. They then use a discriminative model to rerank the translation output using additional non-world level features. The goal in this section is to provide the technical details of their translation approach, which we improve in Section 4. 

\subsection{Word-based Translation Model}

The translation models investigated in \newcite{richardson:17a} use a noisy-channel formulation where $p(\textbf{z} \mid \textbf{x}) \propto  p(\textbf{x} \mid \textbf{z}) p(\textbf{z})$ via Bayes rule. By assuming a uniform prior on output components, $p(\textbf{z})$, the model therefore involves estimating $p(\textbf{x} \mid \textbf{z})$, which under a word-translation model is computed using the following formula: $p( \textbf{x} \mid \textbf{z} ) = \sum_{a \in \mathcal{A}} p(\textbf{x},a \mid \textbf{z})$, where the summation ranges over the set of all many-to-one word alignments $\mathcal{A}$ from $\textbf{x} \to \textbf{z}$, with $|\mathcal{A}|$ equal to $(|\textbf{z} | + 1)^{| \textbf{x} |}$. They investigate various types of sequence-based alignment models \cite{och2003systematic}, and find that the classic IBM Model 1 outperforms more complex word models. This model factors in the following way and assumes an \emph{independent word generation} process:
\begin{equation}
 p(\textbf{x} \mid \textbf{z}) =  \frac{1}{| \mathcal{A} |} \prod_{j=1}^{| \textbf{x} |} \sum_{i=0}^{| \textbf{z} |} p_{t}(x_{j} \mid z_{i})
\end{equation}
where each $p_{t}$ defines a multinomial distribution over a given component term $z$ for all words $x$. 

The decoding problem for the above translation model involves finding the most likely output $\hat z$, which requires solving an $\argmax_{\textbf{z}}$ over Equation 1. In the general case, this problem is known to be $\mathcal{NP}$-complete for the models under consideration \cite{knight1999decoding} largely due to the large space of possible predictions $\textbf{z}$. \newcite{richardson:17a} avoid these issues by exploiting the finiteness of the target component search space (an idea we also pursue here and discuss more below), and describe a constrained decoding algorithm that runs in time $O(|\mathcal{C}|\log |\mathcal{C}|)$. While this works well for small APIs, it becomes less feasible when dealing with large sets of APIs, as in the polyglot case, or with more complex semantic languages typically used in SP \cite{liang2013lambda}.  


\begin{figure*}
\centering
\scriptsize 
\begin{tikzpicture}[scale=.4]
  \node[state,accepting,fill=Gray,thick] (start) [above right=2cm,minimum size=1pt,label=above:{\scriptsize $s_{0}$}]     {$0.0_{0}$};
  \node[state,fill=Gray,thick] (S) [right=1.2cm of start,minimum size=6pt,label=above:{\scriptsize $s_{5}$}]     {$\infty_{5}$};
  \node[state,thick,fill=Gray,minimum size=3pt,label=below:{\scriptsize $s_{1}$}]         (q1) [below=.2cm of S]    {$\infty_{1}$};
  \node[state,thick,fill=Gray,minimum size=3pt,label=above:{\scriptsize $s_{6}$}]         (q2) [right=2cm of S]    {$\infty_{6}$};
    \node[state,thick,fill=Gray,minimum size=5pt,label=below:{\scriptsize $s_{2}$}]         (q10) [right=2cm of q1]     {$\infty_{2}$};
    \node[state,thick,fill=Gray,minimum size=5pt,label=below:{\scriptsize $s_{3}$}]         (q11) [right=2cm of q10] {$\infty_{3}$};
    \node[state,thick,fill=Gray,minimum size=5pt,label=above:{\scriptsize $s_{7}$}]         (q12) [right=2cm of q2] {$\infty_{7}$};
    \node[state,thick,fill=Gray,minimum size=5pt,label=below:{\scriptsize $s_{4}$}]         (q19) [right=2cm of q11] {$\infty_{4}$};
    \node[state,thick,fill=Gray,minimum size=5pt,label=above:{\scriptsize $s_{9}$}]         (q14) [right=2cm of q12] {$\infty_{9}$};
    \node[state,thick,fill=Gray,minimum size=5pt,label=below:{\scriptsize $s_{10}$}]         (q15) [right=1cm of q19] {$\infty_{10}$};
    \node[state,thick,fill=Gray,minimum size=5pt,accepting,label=above:{\scriptsize $s_{11}$}]         (q16) [right=1.2cm of q15] {$\infty_{11}$};
    \node[state,thick,fill=Gray,minimum size=5pt,label=above:{\scriptsize $s_{8}$}]         (q17) [right=1cm of q14] {$\infty_{8}$};
  \path[->] (start)  
             edge[line width=0.7mm,dashed,bend left=10]              node[above] {\footnotesize \texttt{\textbf{2C}}} (S)
             edge[dashed,bend right]              node[below left] {\footnotesize \texttt{\textbf{2Clojure}}} (q1);
  \path[->] (S)  
             edge[line width=0.7mm,bend left=10]              node[above] {\footnotesize \texttt{numeric}}(q2);
   \path[->] (q1)
             edge[thick,bend right=10]              node[below] {\footnotesize \texttt{algo}} (q10);    
   \path[->] (q2)
             edge[line width=0.7mm,bend left=10]              node[above] {\footnotesize \texttt{math}} (q12);
   \path[->] (q10)
             edge[thick,bend right=10]              node[below] {\footnotesize \texttt{math}} (q11);                                
   \path[->] (q11)
             edge[thick,bend right]              node[below] {\footnotesize \texttt{ceil}} (q19)
             edge[thick,bend right]              node[above,pos=0.3] {\footnotesize \texttt{atan2}} (q14); 
   \path[->] (q12)
             edge[thick,bend left=5]              node[above] {\footnotesize \texttt{atan2}} (q14)
             edge[line width=0.7mm,bend left]              node[above,bend left] {\footnotesize \texttt{ceil}} (q17);  
   \path[->] (q19)
             edge[thick,bend right=40]              node[below] {\footnotesize \texttt{x}} (q16);                        
    \path[->] (q14)
             edge[thick,bend left]              node[below] {\footnotesize \texttt{x}} (q15);
    \path[->] (q17)
             edge[line width=0.7mm,bend left]              node[above] {\footnotesize \texttt{arg}} (q16);
    \path[->] (q15)
             edge[thick,bend left]              node[below] {\footnotesize \texttt{y}} (q16);                                            
\end{tikzpicture}

\caption{A \texttt{DAFSA} representation for a portion of the component sequence search space $\mathcal{C}$ that includes math functions in \textbf{C} and \textbf{Clojure}, and an example path/translation (in \textbf{bold}): \textbf{2C} \texttt{numeric math ceil arg}. }
\label{fig:graph}
\end{figure*}
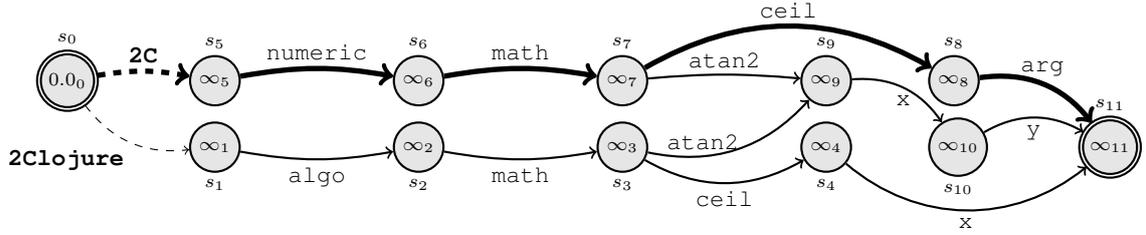

\section{Shortest Path Framework}

To improve the baseline translation approach used previously (Section 3.1), we pursue a graph based approach. Given the formulation above and the finiteness of our prediction space $\mathcal{C}$, our approach exploits the fact that we can represent the complete component search space for any set of APIs as a directed acyclic finite-state automaton (\texttt{DAFSA}), such as the one shown graphically in Figure~\ref{fig:graph}. The underlying graph is constructed by concatenating all of the component representations for each API of interest and applying standard  finite-state construction and minimization techniques \cite{mohri1996some}. Each path in the resulting compact automaton is therefore a well-formed component representation.

Using an idea from \newcite{johnson2016google}, we add to each component representation an \emph{artificial} token that identifies the output programming language or library. For example, the two edges from the initial state $0$ in Figure~\ref{fig:graph} are labeled as \emph{2C} and \emph{2Clojure}, which identify the C and Clojure programming languages respectively. All paths starting from the right of these edges are therefore valid paths in each respective programming language. The paths starting from the initial state $0$, in contrast, correspond to all valid component representations in all languages. 

Decoding reduces to the problem of finding a path for a given text input $\textbf{x}$. For example, given the input \emph{the ceiling of a number}, we would want to find the paths corresponding to the component translations \texttt{numeric math ceil arg} (in C) and \texttt{algo math ceil x} (in Clojure) in the graph shown in Figure~\ref{fig:graph}.   Using the trick above, our setup facilitates both monolingual decoding, i.e., generating components specific to a particular output language (e.g., the C language via the path shown in bold), and polyglot decoding, i.e., generating any output language by starting at the initial state 0 (e.g., C and Clojure). 





We formulate the decoding problem  using a variant of the well-known single source shortest path (SSSP) algorithm  for directed acyclic graphs (\texttt{DAG}s) (\newcite{johnson1977efficient}). This involves a graph $\mathcal{G} = (V,E)$ (nodes $V$ and labeled edges $E$, see graph in Figure~\ref{fig:graph}), and taking an off-line topological sort of the graph's vertices. Using a data structure $d \in \mathbb{R}^{| V |}$ (initialized as $\infty^{| V |}$, as shown in Figure~\ref{fig:graph}), the standard SSSP algorithm (which is the \emph{forward update} variant of the Viterbi algorithm \cite{dynprog}) works by searching forward through the graph in sorted order and finding for each node $v$ an incoming labeled edge $u$, with label $z$,  that solves the following recurrence: 
\begin{equation}
d(v) = \underset{(u,z) : (u,v,z) \in E}{\min}\Big\{d(u) + w(u,v,z) \Big\}
\end{equation}
where $d(u)$ is shortest path score from a unique source node $b$ to the incoming node $u$ (computed recursively) and $w(u,v,z)$ is the weight of the particular labeled edge. The weight of the resulting shortest path is commonly taken to be the sum of the path edge weights as given by $w$, and the output translation is the sequence of labels associated with each edge. This algorithm runs in linear time over the size of the graph's adjacency matrix (\texttt{Adj}) and can be extended to find $k$ SSSPs. In the standard case, a weighting function $w$ is provided by assuming a static weighted graph. In our translation context, we replace $w$ with a translation model, which is used to dynamically generate edge weights during the SSSP search for each input $\textbf{x}$ by scoring the translation between $\textbf{x}$ and each edge label $z$ encountered. 




Given this general framework, many different translation models can be used for scoring. In what follows, we describe two types of decoders based on lexical translation (or unigram) and neural sequence models. Technically, each decoding algorithm involves modifying the standard SSSP search procedure by adding an additional  data structure $s$ to each node (see Figure~\ref{fig:graph}), which is used to store information about translations  (e.g., running lexical translation scores, RNN state information) associated with particular shortest paths. By using these two very different models, we can get insight into the challenges associated with the technical documentation translation task. As we show in Section 6, each model achieves varying levels of success when subjected to a wider range of SP tasks, which reveals differences between our task and other  SP tasks.



\subsection{Lexical Translation Shortest Path} 

\begin{algorithm}[t]
\algnewcommand\algorithmicinput{\textbf{Input:}}
\algnewcommand\INPUT{\item[\algorithmicinput]}
\algnewcommand\algorithmicoutput{\textbf{Output:}}
\algrenewcommand\algorithmicindent{1.5em}
\algnewcommand\OUTPUT{\item[\algorithmicoutput]}
\caption{Lexical Shortest Path Search}
\begin{algorithmic}[1]
\INPUT  \text{Input} $\textbf{x}$ of size $n$, \text{DAG }$\mathcal{G}=(V,E)$, lexical translation function $p_{t}$, source node $b$ with initial score $o$.
\OUTPUT Shortest component path
\State $d[V[\mathcal{G}]] \leftarrow \infty , \pi[V[\mathcal{G}]] \leftarrow Nil, d[b] \leftarrow o  $
\State $s[V[\mathcal{G}],n] \leftarrow 0.0 $\Comment{\textcolor{red}{Shortest path sums at each node}}
\For{each vertex $u \geq b \in V[\mathcal{G}]$ in sorted order }
\For{each vertex and label $(v,z) \in \texttt{Adj}[u]$}
\State \textcolor{red}{$\texttt{score} \leftarrow -log \big[ \thinspace\thinspace \prod_{i}^{n} p_{t}(x_{i} \mid z)+s[u,i]$ \thinspace\thinspace \big]} 
\If{$d[v] >  \texttt{score}$}
\State{$d[v] \leftarrow \texttt{score},\pi[v] \leftarrow u$}
\For {\textcolor{red}{$i$ in $1,..,n$}}\Comment{\textcolor{red}{Update scores}}
\State{\textcolor{red}{$s[v,i] \leftarrow p_{t}(x_{i} \mid z) + s[u,i]$}}
\EndFor
\EndIf
\EndFor
\EndFor
\State \textbf{return} \textsc{FindPath}$(\pi,| V |,b)$
\end{algorithmic}
\end{algorithm}

In our first model, we use the lexical  translation model and probability function $p_{t}$ in Equation 1 as the weighting function, which can be learned efficiently off-line using the EM algorithm.  When attempting to use the SSSP procedure to compute this equation for a given source input $\textbf{x}$, we immediately have the problem that such a computation requires a complete component representation $\textbf{z}$ \cite{finiteMT}.  We use an approximation\footnote{Details about the approx. are provided as supp. material.} that involves ignoring the normalizer $|\mathcal{A}|$ and exploiting the word independence assumption of the model, which allows us to incrementally compute translation scores for individual source words given output translations  corresponding to shortest paths during the SSSP search. 



The full decoding algorithm in shown in Algorithm 1, where the red highlights the adjustments made to the standard SSSP search as presented in \newcite{cormenintroduction}. The main modification involves adding a data structure $s \in  \mathbb{R}^{| V |  \thinspace \times \thinspace | \textbf{x} |}$ (initialized as $0.0^{|V| \times|\textbf{x}|}$ at line 2) that stores a running sum of source word scores given the best translations at each node, which can be used for computing the inner sum in Equation 1. For example, given an input utterance \emph{ceiling function}, $s_{6}$ in Figure~\ref{fig:graph} contains the \emph{independent} translation scores for words  \emph{ceiling} and \emph{function} given the edge label \texttt{numeric} and $p_{t}$. Later on in the search, these scores are used to compute $s_{7}$, which will provide translation scores for each word given the edge sequence \emph{numeric math}. Taking the product over any given $s_{j}$ (as done in line 7 to get \texttt{score}) will give the probability of the shortest path translation at the particular point $j$. Here, the transformation into $-\log$ space is used to find the \emph{minimum} incoming path. Standardly,  the data structure $\pi$ can be used to retrieve the shortest path back to the source node $b$ (done via the \textsc{FindPath} method). 






\subsection{Neural Shortest Path}

Our second set of models use neural networks to compute the weighting function in Equation 2. We use an encoder-decoder model with global attention \cite{bahdanau2014neural,luong2015effective}, which has the following two components: 

\paragraph{Encoder Model}  The first is an \emph{encoder} network, which uses a bi-directional recurrent neural network architecture with LSTM units \cite{hochreiter1997long} to compute a sequence of forward annotations or hidden states $(\overrightarrow{h}_{1},...,\overrightarrow{h}_{| \textbf{x} |})$ and a sequence of backward hidden states $(\overleftarrow{h},...,\overleftarrow{h}_{| \textbf{x} |})$ for the input sequence $(x_{1},...,x_{| \textbf{x}|})$. Standardly, each word is then represented as the concatenation of its forward and backward states: $h_{j} = [\overrightarrow{h}_j, \overleftarrow{h}_j]$. 


\paragraph{Decoder Model} The second component is a \emph{decoder} network, which directly computes the conditional distribution $p(\textbf{z} \mid \textbf{x})$ as follows: 
\begin{align}
p(\textbf{z} \mid \textbf{x}) = \sum_{i=1}^{| \textbf{z} |} \log p_{\Theta}(z_{i} \mid z_{< i}, \textbf{x}) \\
p_{\Theta}(z_{i} \mid z_{< i},\textbf{x}) \sim \texttt{softmax}(f(\Theta,z_{< i},\textbf{x})) 
\end{align} 
where $f$ is a non-linear function that encodes information about the sequence $z_{< i}$ and the input $\textbf{x}$ given the model parameters $\Theta$. We can think of this model as an ordinary recurrent language model  that is additionally conditioned on the input $\textbf{x}$ using information from our encoder. We implement the function $f$ in the following way: 
\begin{align}
f(\Theta,z_{< i},\textbf{x}) = \textbf{W}_{o} \eta_{i} + \textbf{b}_{o} \\
\eta_{i} =  \texttt{MLP}(c_{i},g_{i}) \\
g_{i} = \texttt{LSTM}_{dec}(g_{i-1},\textbf{E}^{out}_{z_{i-1}}, c_{i})
\end{align}
where \texttt{MLP} is a multi-layer perceptron model with a single hidden layer, $\textbf{E}^{out} \in \mathbb{R}^{| \Sigma_{dec} | \times e}$ is a randomly initialized  embedding matrix,  $g_{i}$ is the decoder's hidden state at step $i$, and $c_{i}$ is a context-vector that encodes information about the input $\textbf{x}$ and the encoder annotations. Each context vector $c_{i}$ in turn is a weighted sum of each annotation $h_{j}$ against an attention vector $\alpha_{i,j}$, or $c_{i} = \sum_{j=1}^{| \textbf{x} |} \alpha_{i,j} h_{j}$, which is jointly learned using an additional single layered multi-layer perceptron defined in the following way:
\begin{align}
\alpha_{i,j} \propto \exp(e_{i,j}); \quad  e_{i,j} = \texttt{MLP}(g_{i-1},h_{j}) 
\end{align}


\paragraph{Lexical Bias and Copying} 
In contrast to standard MT tasks, we are dealing with a relatively low-resource setting where the sparseness of the target vocabulary is an issue. For this reason, we experimented with integrating lexical translation scores using a biasing technique from \newcite{arthur2016incorporating}. Their
method is based on the following computation for each token $z_i$: 
\begin{align*}
\texttt{bias}_{i} = \quad\quad\quad\quad\quad\quad\quad\quad\quad\quad\quad\quad\quad\quad\quad\quad \\
	\footnotesize
	\begin{bmatrix}
		p_{t'}(z_{1} \mid x_{1}) & \dots & p_{t'}(z_{1} \mid x_{| \textbf{x} |}) \\
		\vdots & \ddots & \vdots \\
	    p_{t'}(z_{\mid \Sigma_{dec} \mid} \mid x_{1}) &    \dots    & p_{t'}(z_{\mid \Sigma_{dec} \mid} \mid x_{| \textbf{x} |})
	    \end{bmatrix}\begin{bmatrix}
           \alpha_{i,1} \\
           \vdots \\
           \alpha_{i,\mid \textbf{x} \mid}
	\end{bmatrix}
\end{align*}
The first matrix uses the inverse ($p_{t'}$) of the lexical translation function $p_{t}$ already introduced to compute the probability of each word in the target vocabulary $\Sigma_{dec}$ (the columns) with each word in the input $\textbf{x}$ (the rows), which is then weighted by the attention vector from Equation 8. $bias_{i}$ is then used to modify  Equation 5 in the following way:  
\begin{align*}
f_{bias}(\Theta,z_{< i}, \textbf{x}) = {}& \textbf{W}_{o} \eta_{i} + \textbf{b}_{o}  +\\
	         & \log(\texttt{bias}_{i} + \epsilon)
\end{align*}
where $\epsilon$ is a hyper-parameter that helps to preserve numerical stability and  biases more heavily on the lexical model when set lower. 

We also experiment with the \emph{copying} mechanism from \newcite{jia2016data}, which  works by allowing the decoder to choose from a set of latent actions, $a_{j}$, that includes writing target words  according to Equation 5, as done standardly, or copying source words from $\textbf{x}$, or $\texttt{copy}[x_{i}]$ according to the attention scores in Equation 8. A distribution is then computed over these actions using a \texttt{softmax} function and particular actions are chosen accordingly during training and decoding. 

\begin{algorithm}[t]
\algnewcommand\algorithmicinput{\textbf{Input:}}
\algnewcommand\INPUT{\item[\algorithmicinput]}
\algnewcommand\algorithmicoutput{\textbf{Output:}}
\algrenewcommand\algorithmicindent{1.5em}
\algnewcommand\OUTPUT{\item[\algorithmicoutput]}
\caption{Neural Shortest Path Search}
\begin{algorithmic}[1]
\INPUT  \text{Input} $\textbf{x}$, \text{DAG }$\mathcal{G}$, neural parameters $\Theta$ and non-linear function $f$, beam size $l$, source node $b$ with init. score $o$.  
\OUTPUT Shortest component path
\State $d[V[\mathcal{G}]] \leftarrow \infty, d[b] \leftarrow o, \pi[V[\mathcal{G}]] \leftarrow Nil $
\State $s[V[\mathcal{G}]] \leftarrow Nil $\Comment{\textcolor{red}{Path state information}}
\State $s[b] \leftarrow \texttt{InitState()}$\Comment{\textcolor{red}{Initialize source state}}
\For{each vertex $u \geq b \in V[\mathcal{G}]$ in sorted order }
\If{$\texttt{isinf}(d[u])$} \text{continue} \EndIf
\State{$p \leftarrow s[u]$}\Comment{\textcolor{red}{Current state at node $u$, or $z_{< i}$}}
\State{\textcolor{red}{$L^{1}_{[l]} \leftarrow \underset{(v_{1},...,v_{k}) \in \texttt{Adj}[u]}{\argmax}{\texttt{softmax}(f(\Theta,p,\textbf{x})})$}}
\For{each vertex and label $(v,z) \in L$}
\State{\textcolor{red}{\texttt{score} $\leftarrow -\log p_{\Theta}(z \mid p, \textbf{x})+d[u]$}}
\If{$d[v] >  \texttt{score}$}
\State{$d[v] \leftarrow \texttt{score}, \pi[v] \leftarrow u$}
\State{\textcolor{red}{$s[v] \leftarrow \texttt{UpdateState}(p,z)$}}
\EndIf
\EndFor
\EndFor
\State \textbf{return} \textsc{FindPath}$(\pi,| V |,b)$
\end{algorithmic}
\end{algorithm}

\paragraph{Decoding and Learning} The full decoding procedure is shown in Algorithm 2, where the differences with the standard SSSP  are again shown in red. We change the data structure $s$ to contain the decoder's RNN state at each node. We also modify the scoring (line 7, which uses Equation 4) to consider only the top $l$ edges or translations at that point, as opposed to imposing a full search. When $l$ is set to 1, for example, the procedure does a greedy search through the graph, whereas when $l$ is large the procedure is closer to a full search. 

In general terms, the decoder described above works like an ordinary neural decoder  with the difference that each decision (i.e., new target-side word translation) is constrained (in line 7) by the transitions allowed in the underlying graph in order to ensure wellformedness of each component output. Standardly, we optimize these models using stochastic gradient descent with the objective of finding parameters $\hat \Theta$ that minimize the negative conditional log-likelihood of the training dataset.

\subsection{Monolingual vs. Polyglot Decoding} Our framework facilitates both monolingual and polyglot decoding. In the first case, the decoder requires a graph associated with the output semantic language (more details in next section) and a trained translation model. The latter case requires taking the union of all datasets and graphs (with artificial identifier tokens) for a collection of target datasets and training a single model over this global dataset. In this setting, we can then decode to a particular language using the language identifiers or decode without specifying the output language. The main focus in this paper is investigating polyglot decoding, and in particular the effect of training models on multiple datasets when translating to individuals APIs or SP datasets. 

When evaluating our models and building QA applications, it is important to be able to generate the $k$ best translations. This can easily be done in our framework by applying standard $k$ SSSP algorithms \cite{brander1995comparative}. We use an implementation of the algorithm of \newcite{yen1971finding}, which works on top of the SSSP algorithms introduced above by iteratively finding deviating or branching paths from an initial SSSP (more details provided in supplementary materials).

\begin{figure*}
\centering
	\begin{tabular}{c}
		\includegraphics[scale=.47]{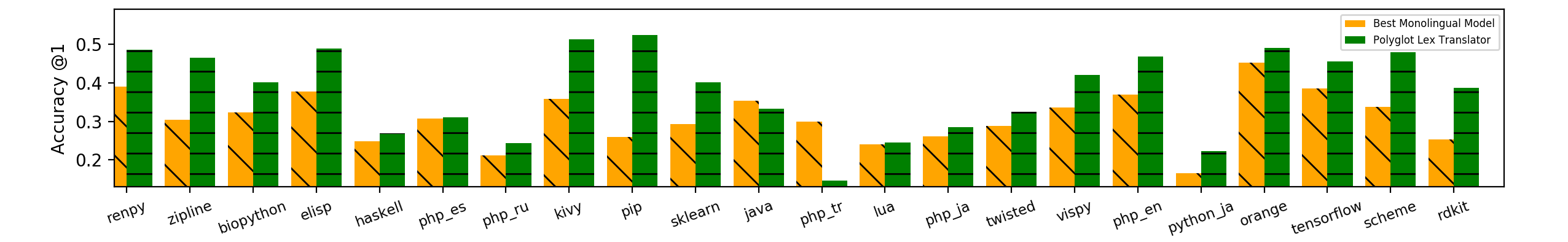} \\[-.23cm]
		\includegraphics[scale=.47]{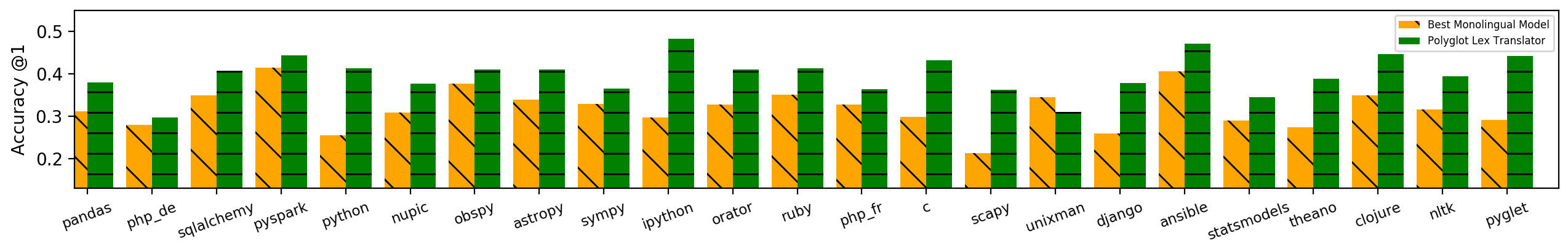}  \\[-.42cm]
	\end{tabular}
\caption{Test \textbf{Acc@1} for the best monolingual models (in yellow/left) compared with the best lexical polyglot model (green/right) across all 45 technical documentation datasets. }
\end{figure*}

\section{Experiments}

We experimented with two main types of resources: 45 API documentation datasets and two multilingual  benchmark SP datasets.  In the former case, our main objective is to test whether training polyglot models (shown as \textbf{polyglot} in Tables 1-2) on multiple datasets leads to an improvement when compared to training individual monolingual models (shown as \textbf{monolingual} in Tables 1-2). Experiments involving the latter datasets are meant to test the applicability of our general graph and polyglot method to related SP tasks, and are also used for comparison against our main technical documentation task.

\subsection{Datasets} 

\paragraph{Technical API Docs} The first dataset includes the Stdlib and Py27 datasets of \newcite{richardson:17a,richardson:17b}, which are publicly available via \newcite{RichardsonData}. Stdlib consists of short description and function signature pairs for 10 programming languages in 7 languages, and Py27 contains the same type of data for 27 popular Python projects in English mined from Github. We also built new datasets from the Japanese translation of the Python 2.7 standard library, as well as the Lua stdlib documentation in a mixture of Russian, Portuguese, German, Spanish and English.

Taken together, these resources consist of 79,885 training pairs, and we experiment with training models on Stdlib and Py27 separately as well as together (shown as \textbf{+ more} in Table 1). We use a BPE subword encoding \citep{sennrich2015neural} of both input and output words to make the representations more similar and transliterated all datasets (excluding Japanese datasets) to an 8-bit latin encoding. Graphs were built by concatenating all function representations into a single word list and compiling this list into a minimized \texttt{DAFSA}. For our global polyglot dataset, this resulted  in a graph with 218,505 nodes, 313,288 edges, and 112,107 paths or component representations over an output vocabulary of 9,324 words. 

\paragraph{Mixed GeoQuery and Sportscaster} We run experiments on the GeoQuery 880 corpus using the splits from \newcite{andreas2013semantic}, which includes geography queries for English, Greek, Thai, and German paired with formal database queries, as well as a seed lexicon or \emph{NP list} for each language. In addition to training models on each individual dataset, we also learn polyglot models trained on all datasets concatenated together. We also created a new mixed language test set that was built by replacing NPs in 803 test examples with one or more NPs from a different language using the NP lists mentioned above (see examples in Figure 4). The goal in the last case is to test our model's ability to handle mixed language input. We also ran monolingual experiments on the English Sportscaster corpus, which contains human generated soccer commentary paired with symbolic meaning representation produced by a simulation of four games.  

For GeoQuery graph construction, we built a single graph for all languages by extracting general rule templates from all representations in the dataset, and exploited additional information and patterns using the Geobase database and the semantic grammars used in \cite{wong2006learning}.  This resulted in a graph with 2,419 nodes, 4,936 edges and 39,482 paths over an output vocabulary of 164. For Sportscaster, we directly translated the semantic grammar provided in \newcite{chenMain} to a \texttt{DAFSA}, which resulted in a graph with 98 nodes, 86 edges and 830 paths. 

\subsection{Experimental Setup}

For the technical datasets, the goal is to see if our model generates correct signature representations from unobserved descriptions using exact match. We follow exactly the experimental setup and data splits from \newcite{richardson:17a}, and measure the accuracy at 1 (\textbf{Acc@1}), accuracy in top 10 (\textbf{Acc@10}), and \textbf{MRR}.  


For the GeoQuery and Sportscaster experiments, the goal is to see if our models can generate correct meaning representations for unseen input. For GeoQuery, we follow \newcite{andreas2013semantic} in evaluating extrinsically by checking that each representation evaluates to the same answer as the gold representation when executed against the Geobase database. For Sportscaster, we evaluate by exact match to a gold representation. 

\begin{table}
\setlength{\arrayrulewidth}{.8pt}
\setlength{\tabcolsep}{5pt}
\centering 
{\scriptsize
	\begin{tabular}{l l l  c  c  c}
		\hline \hline
		&  & \textbf{Method} & \textbf{Acc@1} & \textbf{Acc@10} & \textbf{MRR} \\ \hline
        \multirow{3}{*}{\rotatebox[origin=c]{90}{\parbox[c]{1cm}{\centering \textbf{stdlib}}}} & \textbf{mono.} & RK Trans + \emph{rerank} & 29.9  & 69.2 & 43.1 \\ \cdashline{2-6}
        &  & Lexical SP & \textbf{33.2} & 70.7 & 45.9 \\
        &  \textbf{poly.} & Lexical SP + \emph{more} & 33.1 & 69.7 & 45.5 \\ \cdashline{3-6}
        &  & Neural  SP + \emph{bias} & 12.1 & 34.3 & 19.5 \\ 
        &  & Neural  SP + \emph{copy\_bias} & 13.9 & 36.5 & 21.5 \\ \cline{1-6}
        \multirow{3}{*}{\rotatebox[origin=c]{90}{\parbox[c]{1cm}{\centering \textbf{py27}}}} & \textbf{mono.} & RK Trans + \emph{rerank} & 32.4 & 73.5 & 46.5 \\ \cdashline{2-6}
        &  & Lexical SP & \textbf{41.3} & 77.7 & 54.1 \\
        &  \textbf{poly.} & Lexical SP + \emph{more} & 40.5 & 76.7 & 53.1 \\ \cdashline{3-6}
        &  & Neural  SP + \emph{bias} & 8.7 & 25.5 & 14.2 \\ 
        &  & Neural  SP + \emph{copy\_bias} & 9.0 & 26.9 & 15.1 \\ \cline{1-6}
\end{tabular}}
\label{table:result1}
\caption{Test results on the Stdlib and Py27 tasks averaged over all datasets and compared against the best monolingual results from \newcite{richardson:17a,richardson:17b}, or RK}
\end{table}

\subsection{Implementation and Model Details}

We use the Foma finite-state toolkit of \newcite{hulden2009foma} to construct all graphs used in our experiments.  We also use the Cython version of Dynet \cite{neubig2017dynet} to implement all the neural 
models (see supp. materials for more details). 

In the results tables, we refer to the lexical and neural models introduced in Section 4 as \emph{Lexical Shortest Path} and \emph{Neural Shortest Path}, where models that use copying (\emph{+ copy}) and lexical biasing (\emph{+ bias}) are marked accordingly. We also experimented with adding a discriminative reranker to our lexical models (+ \emph{rerank}), using the approach from \newcite{richardson:17a},  which uses additional lexical (e.g., word match and alignment) features and other phrase-level and syntax features. The goal here is to see if these additional (mostly non-word level) features help improve on the baseline lexical  models. 
\section{Results and Discussion}

\begin{table}
\setlength{\arrayrulewidth}{.8pt}
\setlength{\tabcolsep}{5pt}
\centering 
{\scriptsize
	\begin{tabular}{l l l  c  c  c}
		\hline \hline
		&  & \textbf{Method} & \textbf{Acc@1} & \textbf{Acc@10} & \\ \hline
        & & UBL \cite{Kwiatkowski} & 74.2 & --  \\
        & & TreeTrans \cite{TreeTrans} & 76.8  & -- \\
        & &  nHT \cite{susanto2017semantic} & \textbf{83.3} & --  \\ \cdashline{3-6}
		\multirow{8}{*}{\rotatebox[origin=c]{90}{\parbox[c]{3cm}{\centering \textbf{Standard Geoquery}}}}& \multirow{3}{*}{\rotatebox[origin=c]{90}{\parbox[c]{1cm}{\centering \textbf{monolingual}}}} & Lexical Shortest Path  & 68.6 & 92.4  \\
		& & Lexical Shortest Path + \emph{rerank}         & 74.2 & 94.1  \\ \cdashline{3-6}
        & & Neural Shortest Path                           & 73.5 & 91.1  \\
        & & Neural Shortest Path + \emph{bias}      & 78.0 & 92.8  \\ 
        & & Neural Shortest Path + \emph{copy\_bias}      & 77.8 & 92.1  \\ \cline{2-6}\cline{2-6}
        & \multirow{4}{*}{\rotatebox[origin=c]{90}{\parbox[c]{1.8cm}{\centering \textbf{polyglot}}}} & Lexical Shortest Path                                   & 67.3 & 92.9 \\
        & & Lexical Shortest Path + \emph{rerank}         & 75.2 & 94.7  \\ \cdashline{3-6}
        & & Neural Shortest Path                           & 78.0 & 91.4  \\
        & & Neural Shortest Path + \emph{bias}            & 78.9 & 91.7  \\
        & & Neural Shortest Path + \emph{copy\_bias}      & \textbf{79.6} & 91.9  \\ \cdashline{1-6}
        \multirow{1}{*}{\rotatebox[origin=c]{90}{\parbox[c]{1cm}{\centering \textbf{Mixed}}}} & \multirow{3}{*}{\rotatebox[origin=c]{90}{\parbox[c]{1cm}{\centering \textbf{poly.}}}}  & Best Monolingual Model  & 4.2 & 18.2 \\ \cline{2-6}
        & & Lexical Shortest Path + \emph{rerank}          & 71.1 & 94.3  \\ 
        & & Neural Shortest Path + \emph{copy\_bias}      & \textbf{75.2} & 90.0  \\[.1cm] \cline{1-6}
        &  \multirow{4}{*}{\rotatebox[origin=c]{90}{\parbox[c]{1.5cm}{\centering \textbf{mono.}}}} & PCFG \cite{BB} & 74.2 & --  \\ 
        & & wo-PCFG \cite{BB} & \textbf{86.0}  & --  \\ \cdashline{3-6}
        \multirow{1}{*}{\rotatebox[origin=c]{90}{\parbox[c]{1cm}{\centering \textbf{Sportscaster}}}} & & Lexical Shortest Path & 40.3 & 86.8  \\
        & & Lexical Shortest Path + \emph{rerank} & 70.3 & 90.2  \\ \cdashline{3-6}
        & & Neural Shortest Path  & 81.9 & 94.8  \\ 
        & & Neural Shortest Path + \emph{bias} & 83.4 & 93.9  \\
        & & Neural Shortest Path + \emph{copy\_bias} & 83.3 & 90.5  \\
\end{tabular}}
\caption{Test results for the standard (above) and mixed (middle) GeoQuery tasks averaged over all languages, and results for the English Sportscaster task (below).}
\end{table}

\paragraph{Technical Documentation Results} Table 1 shows the results for Stdlib and Py27. In the monolingual case, we compare against the best performing models in \newcite{richardson:17a,richardson:17b}. As summarized in Figure 3, our experiments show that training polyglot models on multiple datasets can lead to large improvements over training individual models, especially on the Py27 datasets where using a polyglot model resulted in a nearly 9\% average increase in accuracy @1. In both cases, however, the best performing lexical models are those trained only on the datasets they are evaluated on, as opposed to training on all datasets (i.e.,  \emph{+ more}). This is surprising given that training on all datasets doubles the size of the training data, and shows that adding more data does not necessarily boost performance when the additional data is from another distribution. 

The neural models are strongly outperformed by all other models both in the monolingual and polyglot case (only the latter results shown), even when lexical biasing is applied.  While surprising, this is consistent with other studies on \emph{low-resource} neural MT \cite{zoph2016transfer,ostling2017neural},  where  datasets of comparable size to ours (e.g., 1 million  tokens or less) typically fail against classical SMT models. This result has also been found in relation to neural AMR semantic parsing, where similar issues of sparsity are encountered \cite{pengaddressing}. Even by doubling the amount of training data by training on all datasets (results not shown), this did not improve the accuracy, suggesting that much more data is needed (more discussion below). 

Beyond increases in accuracy, our polyglot models support zero-shot translation as shown in Figure 4, which can be used for translating between unobserved language pairs (e.g., ($es$,\texttt{Clojure}), ($ru$,\texttt{Haskell}) as shown in 1-2), or for finding related functionality across different software projects (as shown in 3). These results were obtained by running our decoder model without specifying the output language.  We note, however, that the decoder can be constrained  to selectively translate to any specific programming language or project (e.g., in a QA setting). Future work will further investigate  the decoder's polyglot capabilities, which is currently hard to evaluate since we do not have an annotated set of function equivalences between different APIs.



%
%

\begin{figure*}
\centering 
\scriptsize
\begin{tabular}{| c l l |}
\hline
1. & Source API (stdlib): (\textbf{\emph{es},\textbf{ PHP})} & \textbf{Input}: Devuelve el mensaje asociado al objeto lanzado. \\ \hline
\multirow{3}{*}{\rotatebox[origin=c]{90}{Output}} & \multicolumn{1}{l}{Language: \textbf{PHP}} & Function Translation: \texttt{public string Throwable::getMessage ( void )} \\
& \multicolumn{1}{l}{Language: \textbf{Java}} & Function Translation: \texttt{public String lang.getMessage( void )} \\
&\multicolumn{1}{l}{Language: \textbf{Clojure}} & Function Translation: \texttt{(tools.logging.fatal throwable message \& more)} \\ 
\hline 2. & Source API (stdlib): (\textbf{\emph{ru},\textbf{ PHP})} & \textbf{Input}:{\cyr konvertiruet stroku iz formata} UTF-32\hspace{-.15cm}{\cyr v format} UTF-16. \\ \hline 
\multirow{3}{*}{\rotatebox[origin=c]{90}{Output}}  & \multicolumn{1}{l}{Language: \textbf{PHP}} & Function Translation: \texttt{string PDF\_utf32\_to\_utf16 (  ... )} \\
& \multicolumn{1}{l}{Language: \textbf{Ruby}} & Function Translation: \texttt{String\#toutf16 => string} \\ 
& \multicolumn{1}{l}{Language: \textbf{Haskell}} & Function Translation: \texttt{Encoding.encodeUtf16LE :: Text -> ByteString} \\ 
\hline 3. & Source API (py): (\textbf{\emph{en},\textbf{ stats})} & \textbf{Input: }Compute the Moore-Penrose pseudo-inverse of a matrix. \\ \hline
\multirow{3}{*}{\rotatebox[origin=c]{90}{Output}}  & \multicolumn{1}{l}{Project: \textbf{sympy}} & Function Translation: \texttt{matrices.matrix.base.pinv\_solve( B, ... )} \\ 
& \multicolumn{1}{l}{Project: \textbf{sklearn}} & Function Translation: \texttt{utils.pinvh( a, cond=None,rcond=None,... )} \\ 
& \multicolumn{1}{l}{Project: \textbf{stats}} & Function Translation: \texttt{tools.pinv2( a,cond=None,rcond=None )} \\ 
\hline \hline 4. & Mixed GeoQuery (de/gr) & \textbf{Input}:  Wie hoch liegt der h\"ochstgelegene punkt in  \textgreek{Alampáma}?
 \\ \hline
\multicolumn{3}{| c |}{Logical Form Translation: \texttt{answer(elevation\_1(highest(place(loc\_2(stateid('alabama'))))))}} \\ \hline
\end{tabular}

\caption{Examples of zero-shot translation when running in polyglot mode (1-3, function representations shown in a conventionalized format), and mixed language parsing (4). }
\end{figure*}

\paragraph{Semantic Parsing Results} SP results are summarized in Table 2. In contrast, the neural models, especially those with biasing and copying,  strongly outperform all other models and are competitive with related work. In the GeoQuery case, we compare against two classic grammar-based models, UBL and TreeTrans, as well as a feature rich, neural hybrid tree model (nHT). We also see that the polyglot Geo achieves the best performance, demonstrating that training on multiple datasets helps in this domain as well.  In the Sportscaster case we compare against two PCFG learning approaches, the second, and best performing model (wo-PCFG) involves a grammar model that encodes complex word-order constraints. 

The real advantage of training a polyglot model is shown on the results related to mixed language parsing (i.e., the middle set of results). Here we compared against the best performing monolingual English model (\textbf{Best Mono. Model}), which clearly does not have a way to deal with multilingual NPs. We also find the neural model to be more robust than the lexical models with reranking. 


While the lexical models overall perform poorly on both tasks, the weakness of this model is particularly acute in the Sportscaster case. We found that mistakes are largely related to the ordering of arguments, which these lexical (unigram) models are blind to. That these models still perform reasonably well on the Geo task shows that such ordering issues are less of a factor in this domain. 


\paragraph{Discussion} Having results across related SP tasks allows us to reflect on the nature of the main technical documentation task. Consistent with recent findings \cite{dong2016language}, we show that relatively simple neural sequence models are competitive with,  and in some cases outperform, traditional grammar-based SP methods on benchmark SP tasks. However, this result is not observed in our technical documentation task, in part because this problem is much harder for neural learners given the sparseness of the target data and lack of redundancy. For this reason, we believe our datasets provide new challenges for neural-based SP, and serve as a cautionary tale about the scalability and applicability of commonly used neural models to lower-resource SP problems.



In general, we believe that focusing on polyglot and mixed language decoding is not only of interest to applications  (e.g, mixed language API QA) but also allows for new forms of SP evaluation that are more revealing than only  translation accuracy. When comparing the accuracy of the best monolingual Geo model and the worst performing neural polyglot model, one could mistakingly think that these models have equal abilities, though the polyglot model is much more robust and general.  Moving forward, we hope that our work helps to motivate more diverse evaluations of this type.

\section{Conclusion}

In this paper, we look at learning from multiple API libraries and datasets in the context of learning to translate text to code representations and other SP tasks. To support \emph{polyglot} modeling of this type, we developed a novel graph based decoding method and experimented with various SMT and neural MT models that work in this framework. We report a mixture of positive results specific to each task and set of models using this polyglot modeling idea, some of which reveal interesting limitations of different approaches to SP. We also introduced two new API datasets, and a mixed language version of Geoquery that will be released to facilitate further work on polyglot SP.

\section*{Acknowledgements}
This work was funded by the Deutsche Forschungsgemeinschaft (DFG) via SFB 732, project D2. We thank Alex Fraser for helpful feedback on an earlier draft.  





\bibliography{naaclhlt2018}
\bibliographystyle{acl_natbib}

\end{document}